\documentclass[letterpaper, 10 pt, conference]{ieeeconf}

\IEEEoverridecommandlockouts
\usepackage{amsmath,amsfonts}
\usepackage{textcomp}
\usepackage{stfloats}
\usepackage{graphicx} 
\usepackage{bm}
\usepackage{subcaption}
\usepackage{xcolor}
\usepackage{array}
\usepackage{tikz}

\usepackage{url}
\usepackage{hyperref}
\usepackage{booktabs}
\usepackage{xcolor,colortbl}
\usepackage{cite}
\usepackage[switch]{lineno}
\graphicspath{ }
\title{\LARGE \bf

Proximal powered knee placement: a case study
}

\author{Kyle R. Embry$^{1,2}$, Lorenzo Vianello$^{1}$, Jim Lipsey$^{1}$, Frank Ursetta$^{1}$, Michael Stephens$^{1}$, Zhi Wang$^{1}$,\\ Ann M. Simon$^{1,2}$, Andrea J. Ikeda$^{1}$, Suzanne B. Finucane$^{1}$, Shawana Anarwala$^{1}$, Levi J. Hargrove$^{1,2,3}$
\thanks{$^1$ Center for Bionic Medicine, Shirley Ryan AbilityLab, Chicago, IL, USA $^2$ Department of Physical Medicine and Rehabilitation, Northwestern University, Chicago, IL, USA $^3$ Department of Biomedical Engineering, Northwestern University, Evanston, IL, USA. \newline This work is supported in part the Congressionally Directed Medical
Research Program (CDMRP) Peer Reviewed Orthopedic Research
Program award W81XWH-22-1-0850, and in part by the MSL Renewed Hope Foundation. The contents of this article do not necessarily represent the policy of the funding a, and you should not assume endorsement by the Federal Government.}
}

\begin{document}
\maketitle

%%%%%%%%%%%%%%%%%%%%%%%%%%%%%%%%%%%%%%%%%%%%%%%%%%%%%%%%%%%%%%%%%%%%%%%%%%%%%%
\begin{abstract}
Lower limb amputation affects millions worldwide, leading to impaired mobility, reduced walking speed, and limited participation in daily and social activities. Powered prosthetic knees can partially restore mobility by actively assisting knee joint torque, improving gait symmetry, sit-to-stand transitions, and walking speed. However, added mass from powered components may diminish these benefits, negatively affecting gait mechanics and increasing metabolic cost.Consequently, optimizing mass distribution, rather than simply minimizing total mass, may provide a more effective and practical solution.

In this exploratory study, we evaluated the feasibility of above-knee powertrain placement for a powered prosthetic knee in a small cohort. Compared to below-knee placement, the above-knee configuration demonstrated improved walking speed (+9.2\% for one participant) and cadence (+3.6\%), with mixed effects on gait symmetry. Kinematic measures indicated similar knee range of motion and peak velocity across configurations. Additional testing on ramps and stairs confirmed the robustness of the control strategy across multiple locomotion tasks.

These preliminary findings suggest that above-knee placement is functionally feasible and that careful mass distribution can preserve the benefits of powered assistance while mitigating adverse effects of added weight. Further studies are needed to confirm these trends and guide design and clinical recommendations.

\end{abstract}
%%%%%%%%%%%%%%%%%%%%%%%%%%%%%%%%%%%%%%%%%%%%%%%%%%%%%%%%%%%%%%%%%%%%%%%%%

\section{Introduction}

% lower limb amputation, problematic: reduced quality of life and ability to perform certain activities
Lower limb amputation affects millions of individuals worldwide, leading to substantial impairments in mobility as well as physical and psychological well-being \cite{fanciullacci2021survey}. Reduced walking speed \cite{van2006physical} and overall mobility are associated with decreased participation in social activities and a reduced ability to return to work, which in turn contribute to increased levels of social isolation, depression, and anxiety \cite{calabrese2023hidden}.

% active llp allows some mitigation
Powered prosthetic knees employ actuators to restore knee joint torque by injecting mechanical power into the human–prosthesis–ground system \cite{elery2020design}. Compared with passive prostheses, these devices show considerable promise, with demonstrated benefits including improved ground reaction force symmetry during gait \cite{hunt2022effect}, reduced sit-to-stand transition times \cite{welker2023improving}, and increased walking speeds \cite{song2024continuous}.

% pwr devices augment mass 
However, the integration of powered components increases the mass of these devices compared to microprocessor-controlled and other passive prosthetic knees, which can diminish their functional benefits and hinder widespread user adoption \cite{gehlhar2023review}. Increased prosthesis mass has been consistently associated with adverse gait outcomes, including reduced walking speed \cite{lehmann1998mass}, decreased stance and swing time symmetry \cite{smith2013effects}, and elevated metabolic cost \cite{reginaldi2024effect}. Consequently, restoring joint torque through powered actuation presents an inherent trade-off, whereby the potential benefits of active assistance may be partially offset by the added weight of the prosthesis.

% possible solution: mass placement
%% I decided to use mass placement rather than inertia because:
% inertial has a meaning dependent by the ref system so either we spend some time explaining what we mean with that or we simply remove it
Rather than minimizing total mass, optimizing mass distribution may be a more practical strategy. Prior work shows that distal mass placement substantially degrades gait, reducing walking speed, increasing metabolic cost, and impairing stance- and step-length symmetry in both healthy individuals and transfemoral amputees \cite{ikeda2022impact,browning2007effects,hekmatfard2013effects}, whereas increased proximal mass has minimal impact on gait symmetry in transtibial amputees \cite{seth2020effect}.

\begin{figure}
    \centering
    \includegraphics[width=0.99\linewidth]{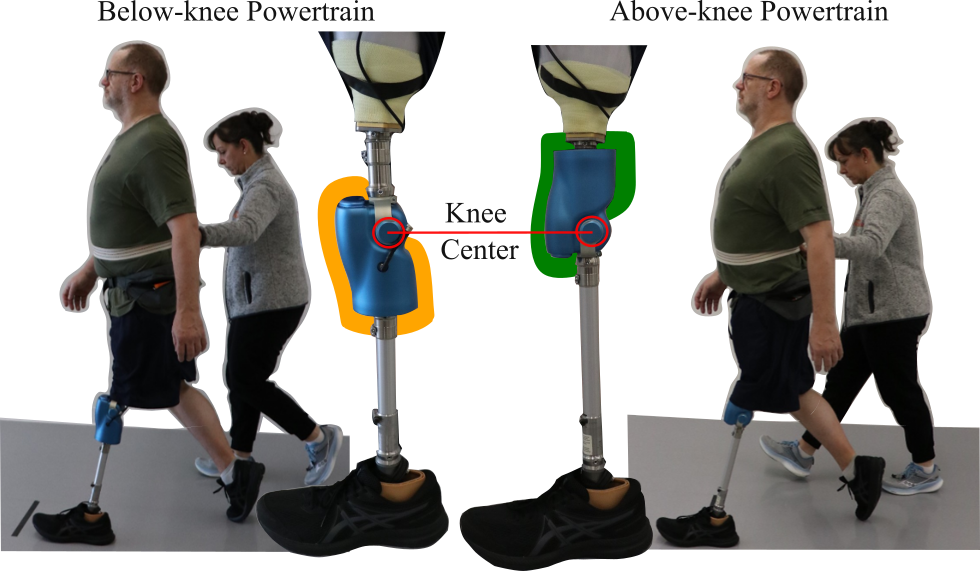}
    \caption{\small{Powered Knee Placement. In this work, we present preliminary example of above knee powertrain placement (right) and compare it to conventional below-knee powertrain placement (left).}}
    \label{fig:initial}
    \vspace{-0.5cm}
\end{figure}

%% Contribution
These findings highlight the potential of optimizing mass placement to enhance the functional benefits of powered prostheses. In this work, we present a case study in which we prototyped a low-profile powered knee (Fig. \ref{fig:initial}) positioned above the knee, thereby shifting mass proximally for the user. The proposed device was evaluated in three individuals with unilateral transfemoral amputation, including one with osseointegration (OI), by comparing gait speed, cadence, gait symmetry, and performance across multiple locomotor tasks.

\section{Methods}
\subsection{Hardware: Powered Knee}

\begin{figure*}
    \centering
    \vspace{0.25cm}
    \includegraphics[width=0.8\linewidth]{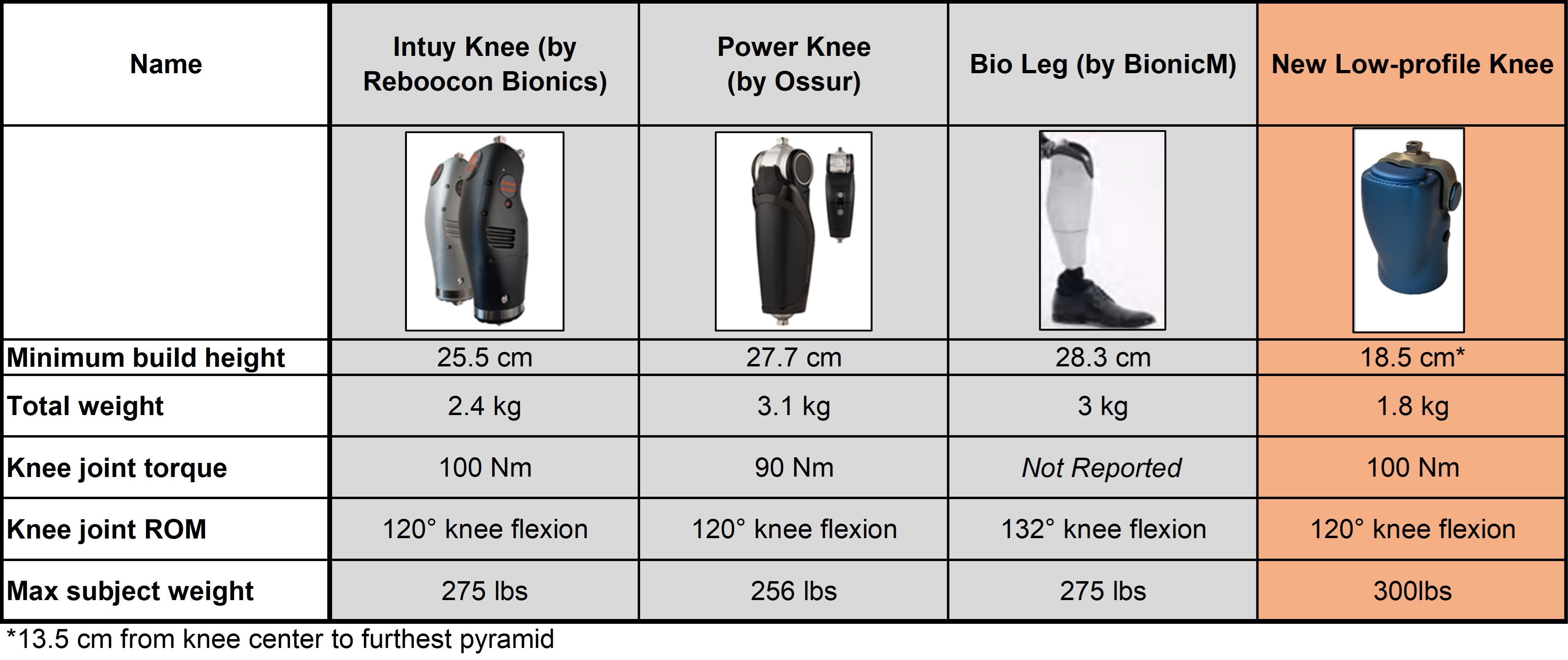}
    \caption{\small{Comparison with commercially available Powered Prosthetic Devices. From left to right, Intuy Knee \cite{elery2020design}, the Össur Power Knee \cite{ossur_power_knee}, the BionicM BioLeg \cite{bionicm_bioleg}, and our prototype.}}
    \label{fig:devices}
\end{figure*}

% The device
Due to the substantial axial length of powered knee powertrains, implementing an above-knee powertrain has historically been impractical. Existing powered knee prototypes and commercially available devices generally exhibit a build height of approximately 24 cm between the most proximal and distal pyramid adapters \cite{elery2020design, simon2022ambulation, ossur_power_knee, azocar2018design, tran2022lightweight}, as illustrated in Fig.~\ref{fig:devices}. In contrast, the proposed low-profile powered knee prototype has a total build height of 18.5 cm as shown in Fig.~\ref{fig:devices}, with only 13.5 cm between the knee center and most distal pyramid adaptor. This reduced profile enables, for the first time, a systematic exploration of both above- and below-knee powertrain placement across a wide range of residual limb lengths.

The prototype has a total mass of 1.8 kg and is capable of delivering knee joint torques of up to 100 Nm. The device integrates a two degree-of-freedom load cell (operating at 100 Hz), a joint encoder to measure position
(250 Hz), a motor encoder to measure velocity (250 Hz), motor current (250 Hz) and an inertial measurement unit (250 Hz).

\subsection{Hierarchical Impedance Control}
\label{sec:ctrl}

A three-level hierarchical software control structure was used to operate the device \cite{simon2022ambulation}. The high-level controller identifies the activity to be performed, namely, overground walking, ramp ascent/descent, stair ascent/descent, or sit/stand transitions triggered via a key fob operated by the prosthetist. At the mid level, a state machine governs transitions between the different gait phases, based on signals from embedded sensors, specifically the load-cell vertical force and inertial measurement unit (IMU) anglular measurements. Finally, at the lowest level of the software stack, a joint-level impedance controller regulates knee motion according to the following equation::

\begin{equation}
    \tau = - k(\theta - \theta^{eq})  - b \dot{\theta},
\end{equation}

\noindent where $\tau$ represents the joint torque, $\theta$ and $\dot{\theta}$ represent the knee angle and velocity, and $k, b$, and $\theta^{eq}$ denote the stiffness, damping coefficient, and equilibrium angle. A subset of these parameters can be tuned for a user.

While the controller present the same structure both for the above- and below- knee powertrain placement two major consideration are necessary.  
\subsubsection{IMU Sensors} 
 Since the IMU is physically integrated into the main body of the prosthesis, its measurements differ between configurations. In the above-knee powertrain placement, the IMU no longer measures shank orientation as in the below-knee configuration; instead, it directly assesses the thigh angle.
This requires modifications to the controller to ensure correct transitions between gait phases. Specifically, if $\theta_{\mathrm{IMU}}$ denotes the sagittal angle measured by the IMU and $q$ is the joint encoder reading, the shank and thigh angles for the below-knee placement are given by:
\begin{align}
    \theta_{shank} &= \theta_{IMU}\\
    \theta_{thigh} &= \theta_{shank} - q
\end{align}
while for the above knee powertrain placement:
\begin{align}
    \theta_{shank} &= \theta_{thigh} + q\\
    \theta_{thigh} &= \theta_{IMU} - 180^{\circ}.
\end{align}

\subsubsection{Load cell}  In the below-knee placement, the load cell measures the forces transferred from the prosthesis to the ground. In the above-knee configuration, the sensor instead measures the forces between the knee and the socket, or between the knee and the OI interface. 
While this does not require major changes to the controller design, it affects the load cell readings, necessitating ad hoc calibration of load thresholds to transition between states, and preventing direct comparison between placements. A more detailed discussion of this aspect is provided in Sec.~\ref{sec:res:kin}.

\subsection{Experimental Setup}

We tested the above knee prosthetic placement with three unilateral transfemoral prosthesis users. All 3 participants were fitted with an Ottobock Trias Foot. This study was approved by Northwestern University Internal Review Board  (IRB NUMBERS: STU00209522 and STU00215805) and participants provided written informed consent. Table. \ref{tab:participants} displays individual patients demographics.  

To fit the above-knee prosthesis, we selected patients with sufficient space between the residual limb and the knee center of rotation to accommodate the device (at least 13.5cm). Since no previous studies have tested a similar approach, we based our evaluation on individual fitting and testing. Additional considerations regarding the fitting are discussed in Sec. \ref{sec:discussion}.

Across the three participants, two (TF01 and TF02) experienced the powered knee in both the below-knee and above-knee configurations. For both configurations, a preliminary phase of control parameter tuning was performed.

After minimal accommodation time, these two participants completed three trials of overground walking on a pressure-sensitive walkway (GAITRite$^{\circledR}$, USA), which allowed measurement of foot placement and temporal gait parameters. Participant TF01 performed the trials at both self-selected and fast walking speeds, whereas participant TF02 performed the trials at a self-selected speed only. 
Each walking condition was repeated three times.

Participant TF03 performed additional activities to validate the ability of both the prosthetic device and the controller to accommodate multiple locomotion modes, including ramp and stair ascent and descent in the above-knee powertrain condition. Specifically, the participant completed seven ramp ascent and descent trials, three stair ascent and descent trials, and twenty 10-m overground walking trials. Stair trials were performed using a reciprocal stair-climbing strategy and were completed with minimal training or acclimation to the above-knee powertrain configuration.

\begin{table}[t]
    \centering
        \vspace{0.4cm}
    \begin{tabular}{|c|c|c|c|c|c|c|c|c|}
        \hline
         \rotatebox{90}{Subject ID} & \rotatebox{90}{Age [yrs]} & \rotatebox{90}{\shortstack{Time since\\amputation [y]}} & \rotatebox{90}{Gender}  & \rotatebox{90}{Etiology} & \rotatebox{90}{Height [cm]} & \rotatebox{90}{Weight [kg]} & \rotatebox{90}{K-level} &  \rotatebox{90}{Osseointegrated }\\
       \hline
       \hline
        TF01 & 49.9 & 47.4 & M  & Trauma & 190.5 & 106.1 & K4 & N\\ % TF50
        TF02 & 31.6 & 16.2 & M  & Sarcoma & 171.5 & 88.6 & K3 & Y \\ % TF90
        TF03 & 50.4 & 1.6 & M  & Trauma & 182.9 & 87.3 & K3 & Y\\ %TF92
        \hline
    \end{tabular}
    \caption{\small{Participants demographics: From left to right: the Subject Identifier; the Age at intake, namely the participant’s age at the time of enrollment in the study; Time since amputation. K-level refers to the Medicare Functional Classification Level, indicating the participant’s mobility potential (K3 = community ambulator with variable cadence, K4 = active adult or athlete). \textit{Osseointegrated} indicates whether the patient underwent osseointegration and uses an osseointegrated (OI) interface rather than a conventional socket.}}
    \label{tab:participants}
\end{table}

\subsection{Data Analysis}

For all participants, we collected the following sensor data embedded in the device: a two-degree-of-freedom load cell to measure internal forces and moments, knee encoders to measure knee angle and angular velocity, motor current sensors, and an IMU to measure shank and thigh orientations. For participants TF01 and TF02, these data were used to compare kinematic and dynamic profiles between the above-knee and below-knee powertrain placement. For participant TF03, the data were used to illustrate representative sensor signals and commanded kinematic profiles across multiple activities, including overground walking and ramp and stair ascent and descent. 

From the instrumented walkway, we measured foot placement and temporal gait parameters, which were used to compute walking speed and cadence. Additionally, spatiotemporal measures were computed for each limb, including step time, step length, swing and stance durations and their percentages of the gait cycle, and step width. For each measure, computed separately for the left and right limbs, the average ($x_l$, $x_r$) across the $8$ meters of the pressure-sensitive walkway was calculated after removing the first and last steps. At this point gait symmetry was quantified as:

\begin{equation}
    SI = \frac{min(x_l, x_r)}{max(x_l, x_r)}.
\end{equation}

This dimensionless measure ($SI \in [0,1]$) quantifies the degree of gait symmetry during locomotion, with higher values indicating greater symmetry. For all computed measures, we report both the inter-subject mean and standard deviation (mean ± SD) across the three walking iterations on the walkway, as well as the across-subject mean and standard deviation.

\section{Results}

\subsection{Analysis Kinematics and Dynamics}
\label{sec:res:kin}

\begin{figure}[t]
    \centering
        \vspace{0.25cm}
    \includegraphics[width=0.99\linewidth]{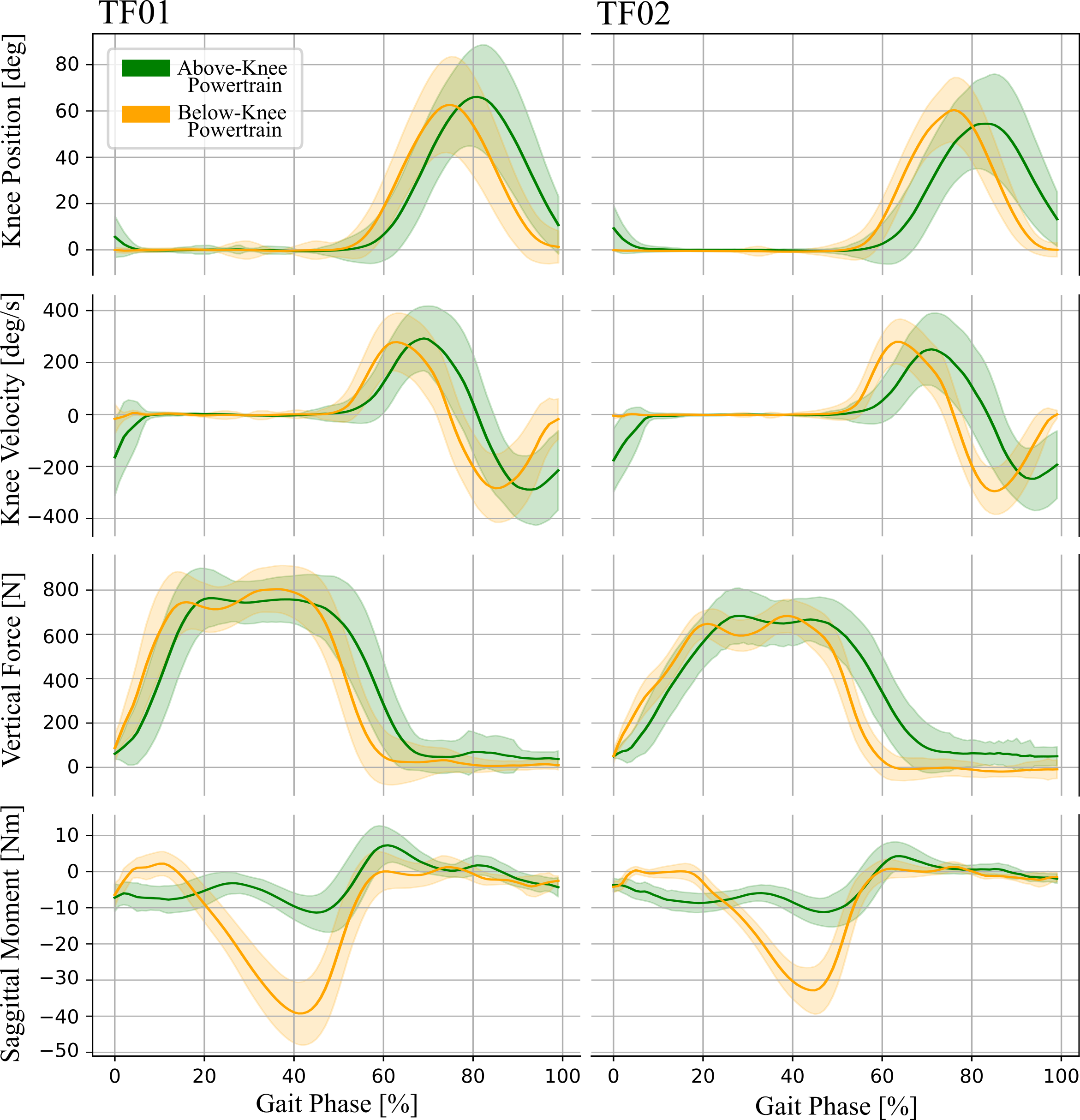}
    \caption{\small Comparison of Kinematic during level-ground walking for Above-Knee (green) and Below-knee (orange) powertrain placement for both TF01 (left) and TF02 (right).The plot displays the mean (solid line) and the $\pm$ standard deviation region, shown as a shaded area, across all strides. From top to bottom: Knee position and velocity measured by knee encoder, Vertical force and saggital moment measured by the load-cell. On the x-axis the gait phase in percent between two consecutive heelstrikes. }
    \label{fig:kneeAngle}
\end{figure}

Onboard prosthetic measurements are displayed in Fig.~\ref{fig:kneeAngle}. The analysis of knee kinematics shows generally comparable behavior in both knee angle (first row) and knee angular velocity (second row) across the two participants. In particular, the knee range of motion (in deg) is of similar magnitude between configurations (TF01 Above-Knee 75.8~deg, TF01 Below-Knee 65.9~deg, TF02 Above-Knee 63.0~deg, TF02 Below-Knee 63.3~deg). Likewise, the peak knee angular velocity (in deg/s) remains comparable across conditions (TF01 Above-Knee 345~deg/s, TF01 Below-Knee 299~deg/s, TF02 Above-Knee 315~deg/s, TF02 Below-Knee 305~deg/s).

The third row of Fig.~\ref{fig:kneeAngle} displays the vertical forces measured by the load cell. As discussed in Sec.~\ref{sec:ctrl}, this sensor signal has a slightly different physical interpretation in the above-knee configuration, where it measures the interaction force between the socket and the prosthetic device, whereas in the below-knee configuration it reflects the interaction forces between the device and the ground. Nevertheless, during overground walking this sensor signal does not appear to be substantially affected by this difference. In fact, during the stance phase, in both configurations, the measured forces fall within the range expected to support the user’s body weight, while during the swing phase the measured forces are close to zero.

The last row of Fig.~\ref{fig:kneeAngle} displays the sagittal moments measured by the load cell. From visual inspection, it is evident that during the stance phase the measured moment is substantially higher in the below-knee configuration (TF01 Above-Knee $-14$~Nm, TF01 Below-Knee $-41$~Nm, TF02 Above-Knee $-12$~Nm, TF02 Below-Knee $-34$~Nm). This difference is primarily due to the distinct sensor placement, which results in a longer effective lever arm during stance in the below-knee configuration, where the kinematic chain originates at the prosthetic foot–ground contact. 
While this difference in sensor readings did not affect the correct operation of the state machine, it prevents a direct and fair comparison of the ground reaction moments generated by the user across configurations.

\subsection{Walking speed and Cadence}

\begin{figure}[t]
    \centering
    \vspace{0.25cm}
    \includegraphics[width=0.99\linewidth]{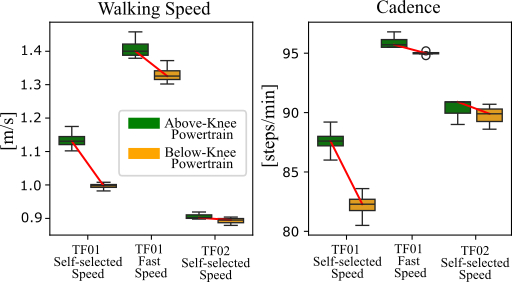}
    \caption{\small Walking Speed (m/s) and Cadence (steps/min) for each participant and condition (Self-selected and Fast speed), measured using the pressure-sensitive walkway. Above-Knee placement is shown in green, and Below-Knee placement in orange. For both walking speed and cadence, higher values indicate respectivelly higher walking speed and cadence.}
    \label{fig:velAndCad}
\end{figure}

Analysis of walking speed and cadence is shown in Fig. \ref{fig:velAndCad}. Results suggest a faster velocity with the above-knee configuration (TF01 self-selected: $1.135 \pm 0.03$, TF01 fast-speed: $1.410 \pm 0.035$, TF02 self-selected: $0.906 \pm 0.0111$) compared to the below-knee placement (TF01 self-selected: $0.996 \pm 1.0$, TF01 fast-speed: $1.331 \pm 0.030$, TF02 self-selected: $0.892 \pm 0.012$), resulting in an improvement of 9.2\% for TF01 and 1.5\% for TF02.

Similar results have been observed for cadence, namely a higher number of steps per minute using the above-knee configuration (TF01 self-selected: $87.6 \pm 1.3$, TF01 fast-speed: $95.9 \pm 0.61$, TF02 self-selected: $90.2 \pm 1.0$) compared to the below-knee placement (TF01 self-selected: $82.2 \pm 1.3$, TF01 fast-speed: $95.0 \pm 0.2$, TF02 self-selected: $89.7 \pm 1.05$), resulting in an improvement of 3.6\% for TF01 and 0.6\% for TF02.

\subsection{Gait symmetries}

\begin{figure}[t]
    \centering
    \vspace{0.25cm}
    \includegraphics[width=0.85\linewidth]{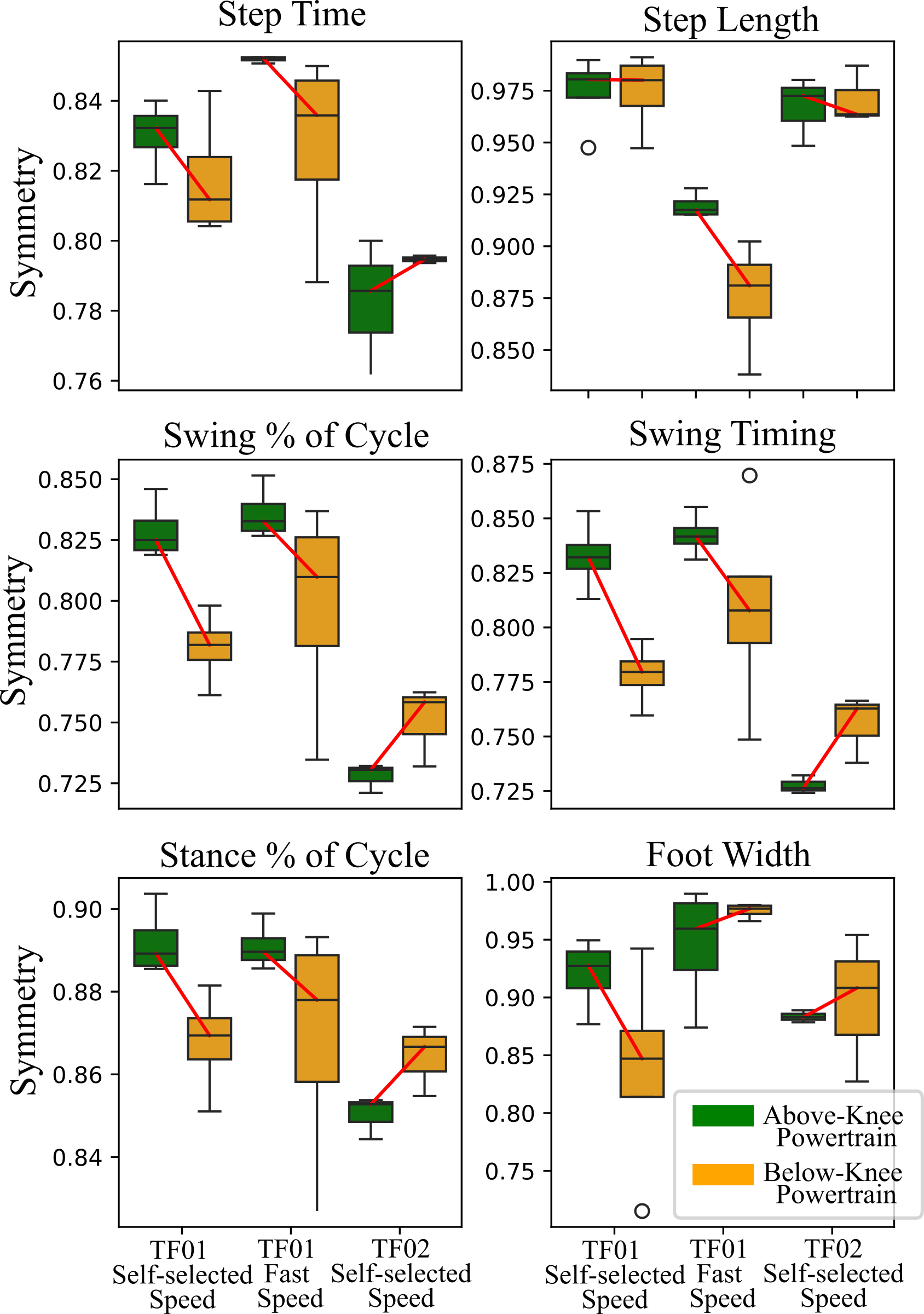}
    \caption{\small Gait symmetries measured with pressure-sensitive walkway for participants TF01 and TF02: Above-Knee configuration is shown in green, and Below-Knee prosthetic placement in orange. From top to bottom and left to right, the plots display symmetries for: step time, step length, swing phase percentage, stance phase percentage, and step width. Higher values indicate more symmetric behavior between the two legs.}
    \label{fig:symmetries}
\end{figure}

Results related to gait symmetries are displayed in Fig.~\ref{fig:symmetries}. Gait symmetries generally show an improvement when transitioning from the below-knee to the above-knee powertrain placement (i.e., values closer to one) for participant TF01, both at self-selected and fast speeds (Step Time +2.24\%, Step Length +2.36\%, Swing percentage +5.45\%, Stance percentage +2.64\%, Step width +2.93\%). 

Conversely, results for TF02 show a negative trend when transitioning from the below-knee to the above-knee powertrain placement (Step Time $-1.53$\%, Step Length $-0.41$\%, Swing percentage $-3.05$\%, Stance percentage $-1.62$\%, Step width $-1.46$\%).

\subsection{Validation on multiple activities}

\begin{figure*}[t]
    \centering
    \vspace{0.25cm}
    \includegraphics[width=0.9\linewidth]{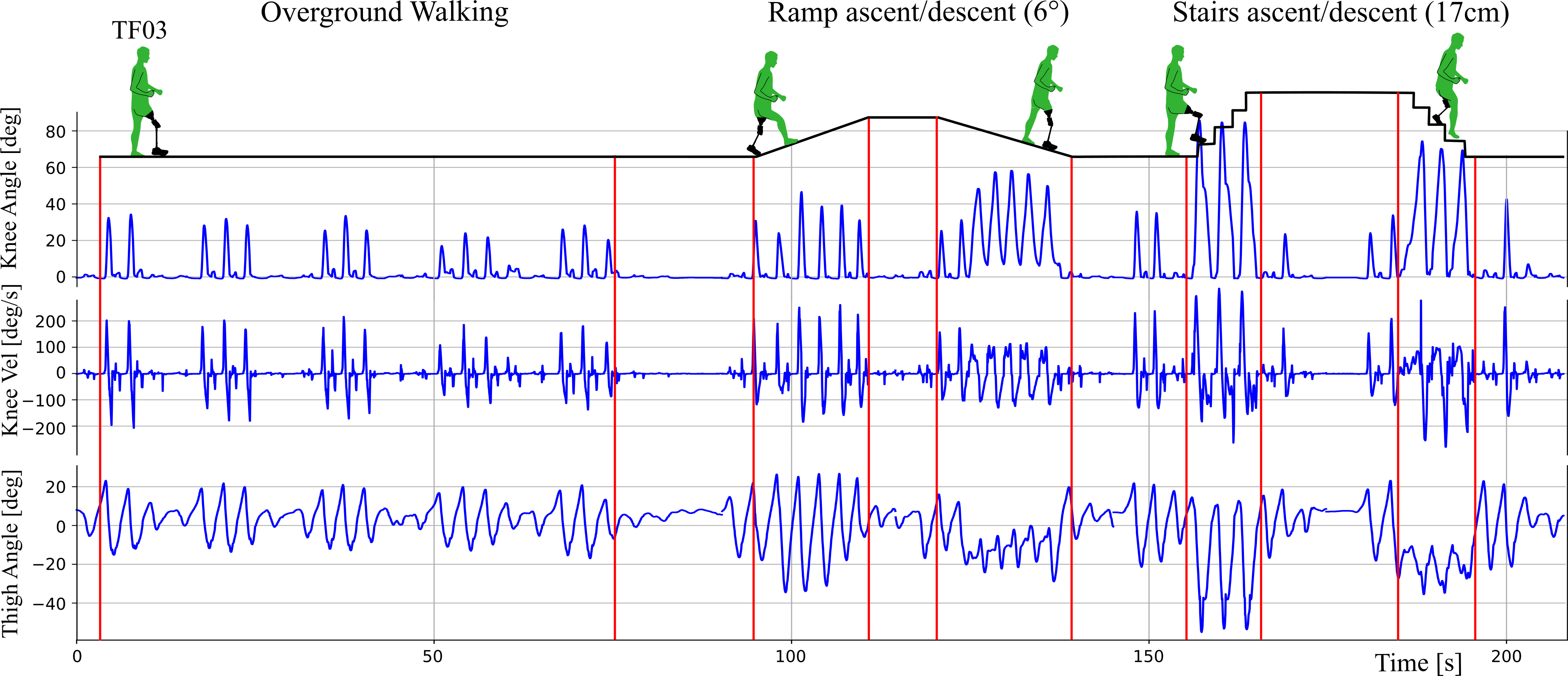}
    \caption{\small Subject TF03 performed multiple activities while wearing the prosthetic device in Above-Knee configuration. Here we display a representative example of the user joint kinematic while performing ramp ascending and descending as well as stairs ascending and descending.  }
    \label{fig:multipleAct}
\end{figure*}

Fig.~\ref{fig:multipleAct} showcases the ability of participant TF03 to perform multiple locomotion activities while wearing the powered leg in the above-knee powertrain placement. The results are consistent with expected behavior, showing an increased knee range of motion compared to overground walking (approximately $30^\circ$). During ramp negotiation, the knee range of motion increased to approximately $40^\circ$ during ramp ascent and $45^\circ$ during ramp descent. The largest excursions were observed during stair ambulation, reaching approximately $80^\circ$ during stair ascent and $70^\circ$ during stair descent.

\section{Discussion}
\label{sec:discussion}

To the best of our knowledge, this study presents the first experimental validation of an above-knee powertrain prosthesis placement. This exploratory work demonstrates that such a configuration is feasible, potentially extending the applicability of existing powered prosthetic platforms. The above-knee powertrain was equally successful during overground walking and provided powered assistance during energy-consuming activities, such as stair ascent. Moreover, the same control strategies and many identical parameter settings were used across multiple ambulation modes for both the above- and below-knee powertrain conditions.

% consideration about the amputation
At the same time, the proposed approach requires a more thorough investigation of fitting constraints and participant inclusion criteria. Residual limb length varies substantially across individuals, as amputation procedures are highly heterogeneous. Although the American Academy of Orthopaedic Surgeons recommends transection of the femur 9–17~cm above the knee center \cite{michael2004atlas, azar2020campbell, baum2008correlation}, residual limbs are often preserved as long as possible to promote soft-tissue health and improve socket control. The use of an above-knee powered prosthesis on a long residual limb would create a knee height difference between the prosthetic and sound knee. Further work is required to determine the longest relative residual limb length that can benefit from an above-knee powered prosthesis \cite{ramakrishnan2017effect}. 
Transfemoral amputees with shorter residual limb lengths may gain greater functional mobility benefits from powered prostheses, as they rely more heavily on prosthetic assistance due to reduced biological leverage. Positioning more of the device weight proximally may improve weight tolerance by reducing swing-phase inertia and hip loading, thereby facilitating these benefits.
Furthermore, the approach may be particularly advantageous for individuals with osseointegrated prostheses, where the absence of a traditional socket reduces constraints associated with socket fit, thereby allowing greater flexibility in device design and mass distribution. %While still preliminary, the results of this study could inform future discussions on femoral transection strategies, potentially guiding reconsideration of optimal residual limb length to maximize compatibility with above-knee powered prostheses, though further research is required before any clinical recommendations can be made.

% consideration regarding the kinematics and dyn
In particular, onboard kinematic measurements suggest comparable knee kinematics between above-knee and below-knee configurations, supporting the feasibility of the proposed placement. Onboard dynamic measures, like load-cell measurements, however, require careful interpretation due to differences in sensor location between configurations. Future studies should aim to harmonize force and moment measurements, for example by relocating the load cell to the shank or ankle. While this may further reduce the size of the above-knee prosthetic structure, it would also introduce changes in mass distribution that must be considered. It is also possible that the reduced measured sagittal moments partially reflect a true reduction in the moments perceived by the user.

% Considerations regarding walking speed, cadence and symmetry
Consistent with this interpretation, results related to walking speed, cadence, and gait symmetry indicate a trend toward increased walking speed and cadence for both participants, while symmetry improvements were observed in only one participant. These observations warrant further investigation with longer accommodation times in adequately powered studies, incorporating kinetic and metabolic outcome measures as well as subjective evaluation.
%% multiple activities
Finally, the proposed platform was successfully tested during additional locomotion tasks, demonstrating the robustness of the control strategy and its adaptability to the above-knee configuration. Further experimentation and validation are expected to refine the tuning of the state-machine parameters and to support clinical translation of this novel approach.

\section{Conclusion}
This study provides the first validation of an above-knee placement for a powered prosthetic powertrain. The results demonstrate functional feasibility with minimal control modifications, highlighting the potential for broader deployment across different prosthetic devices.
Compared to below-knee placement, the above-knee configuration yielded higher walking speed and cadence, with participant-dependent effects on gait symmetry. Kinematic analyses showed comparable knee range of motion and peak velocity between configurations for two participants.
The platform also proved robust across multiple locomotion tasks for one participant, including ramp and stair ambulation. Future studies with larger cohorts, additional kinetic and metabolic metrics, and systematic optimization of fitting criteria are required to validate these preliminary findings and support clinical translation.

\bibliographystyle{IEEEtran}
\bibliography{IEEEabrv,main}

\end{document}